\documentclass{article}

\usepackage[preprint]{nips}

\usepackage[utf8]{inputenc} % allow utf-8 input
\usepackage[T1]{fontenc}    % use 8-bit T1 fonts
\usepackage{hyperref}       % hyperlinks
\usepackage{url}            % simple URL typesetting
\usepackage{booktabs}       % professional-quality tables
\usepackage{amsfonts}       % blackboard math symbols
\usepackage{nicefrac}       % compact symbols for 1/2, etc.
\usepackage{microtype}      % microtypography
\usepackage{xcolor}         % colors
\usepackage{graphicx} 
\usepackage{amsmath}
\usepackage{multirow}
\usepackage{booktabs}
\usepackage{amsmath}

\usepackage{natbib}
\newcommand{\scoregain}[2]{%
  \(\begin{array}{@{}c@{}}
      #1\\[-1.5pt]
      \scriptstyle #2
    \end{array}\)}
\title{Distill What the Student Can See: Fisher-Projected On-Policy Distillation for Vision-Language Models}

\author{%
  Leyan Xue\\
  College of Intelligence and Computing\\ Tianjin University\\ Tianjin, China\\
  \And
  Feng Xiong 
  \thanks{equal contribution}
  \\
  College of Intelligence and Computing\\ Tianjin University\\ Tianjin, China\\
  \And
  Mingjun Ma \\
  College of Intelligence and Computing\\ Tianjin University\\ Tianjin, China\\
  \And
  Changqing Zhang 
  \thanks{Corresponding author.}
  \\
  College of Intelligence and Computing\\ Tianjin University\\ Tianjin, China\\
   \\
}

\begin{document}

\maketitle

\begin{abstract}
  On-policy distillation (OPD) samples trajectories from the current student policy and minimizes token-level divergence between student and teacher next-token distributions at prefixes along those trajectories. This aligns the distillation states with the student's own generation distribution. However, it still assumes that the complete teacher distribution is an appropriate target across student capacities. In vision--language reasoning, teacher corrections can depend on visual distinctions that a compact student cannot represent. Our target-scaling study shows that, as the target approaches the complete teacher distribution, the student realizes less of the prescribed shift and obtains worse downstream performance. We therefore propose \emph{Fisher-Projected On-Policy Distillation} (FP-OPD), which distills only locally realizable teacher corrections. FP-OPD uses continuous visual perturbations to estimate the student's local visual tangent space and projects the centered teacher--student log-probability gap onto this space under the student's Fisher metric. The resulting capacity-aware target is optimized with full-vocabulary reverse KL on student trajectories, retaining the standard OPD framework. In 8B-to-2B distillation, FP-OPD improves all seven evaluated multimodal benchmarks. It raises the average score by 2.77 points over the pretrained student and by 1.60 points over standard OPD. These results demonstrate that locally realizable teacher corrections provide a more effective target for distilling compact vision--language models.
\end{abstract}

% Introduction for FP-OPD v6.
% Required bibliography keys:
% hinton2015distilling, agarwal2024gkd,
% lu2025onpolicydistillation, amari1998natural,
% xu2025speculative, zhang2025capacitygap,
% qian2025goodteacher.

\section{Introduction}
\label{sec:introduction}

\begin{figure*}[t]
    \centering
    \includegraphics[width=\textwidth]{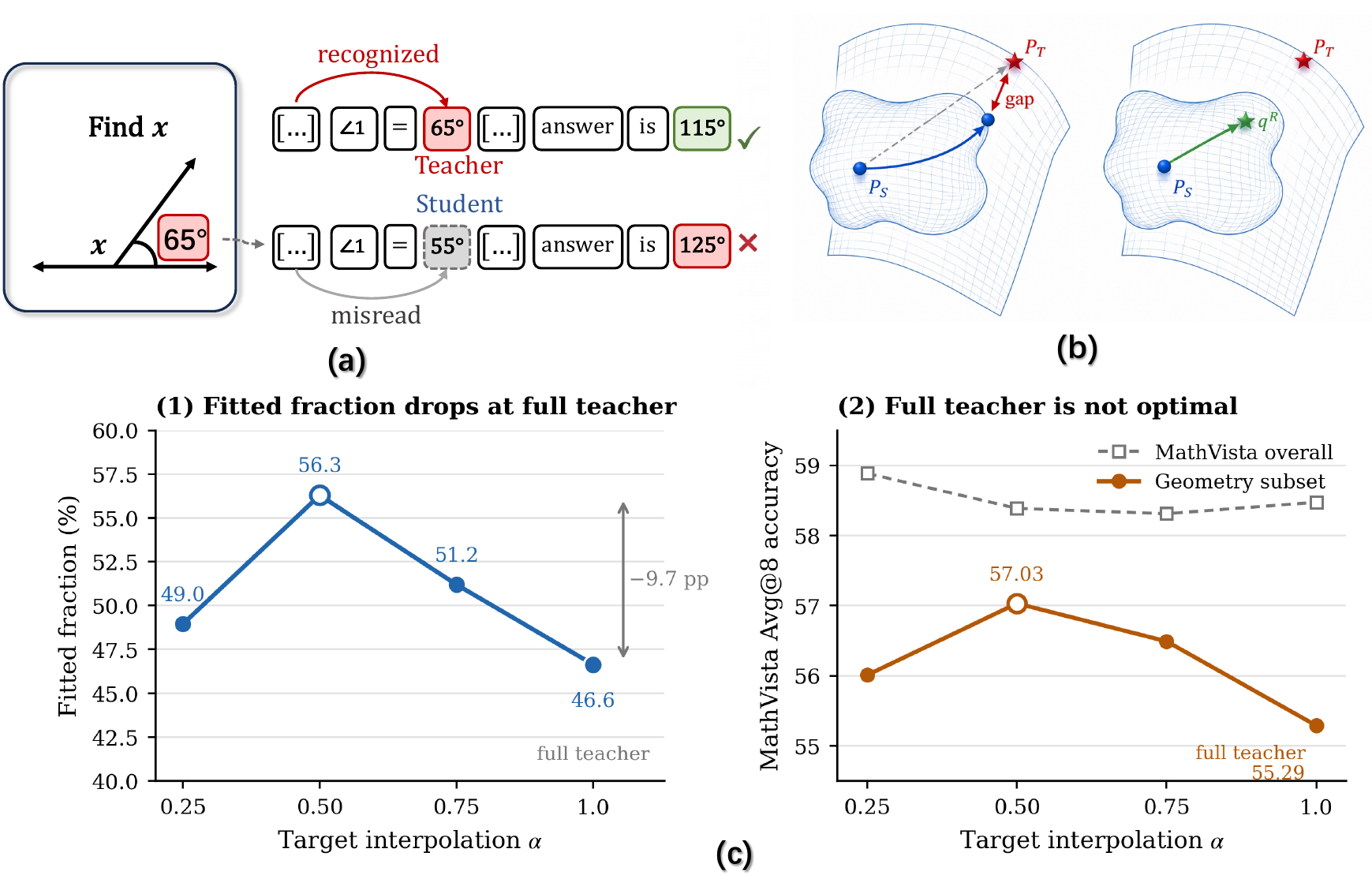}
    \caption{\textbf{Motivation and empirical evidence for capacity-aware
    on-policy distillation.}
    (a) Even on a shared student-generated prefix, teacher supervision can
    reflect visual evidence that is not represented correctly by the
    student, leading to different token predictions and final answers.
    (b) The complete teacher target $p_T$ can lie outside the student's local
    response geometry, whereas Fisher projection produces a reachable target
    $q^R$ within the student-supported region.
    (c) After one epoch of 8B-to-2B OPD training, the full teacher target
    ($\alpha=1$) is harder to fit than an intermediate target and yields
    lower performance on the MathVista Geometry subset.}
    \label{fig:motivation_overview}
\end{figure*}

Knowledge distillation conventionally trains a compact student to match the
predictive distribution of a larger teacher
\citep{hinton2015distilling}. For autoregressive reasoning models, training
only on fixed teacher trajectories creates a mismatch between the prefixes
seen during distillation and the states reached by the student at inference
\citep{ICLR2025_a2747a38}.
On-policy distillation (OPD) instead samples student trajectories and matches
student and teacher next-token distributions at their prefixes, aligning
distillation with student-visited states
\citep{agarwal2024policy,lu2025onpolicydistillation}. In vision-language
reasoning, aligning the visited prefixes resolves where
supervision is applied but leaves a second mismatch in what supervision
requests. Teacher corrections can depend on visual evidence absent from the
compact student's representation. Sharing the rollout therefore does not
make the complete teacher distribution compatible with the student's
capacity. Prior distillation studies likewise show that increasing teacher
capacity does not necessarily improve a compact student when the capacity
gap is large \citep{zhang-etal-2025-towards-law,Qian_2025_ICCV}. Here,
standard OPD minimizes token-level reverse KL to the complete teacher
distribution, treating every correction as student-expressible.

An important source of this mismatch in VLMs is how models represent the
same visual input. Models with different capacities can form different visual
representations and extract different evidence, which changes the next-token
distribution even at a shared prefix.
In Figure~\ref{fig:motivation_overview}(a), the teacher identifies the marked
angle as $65^\circ$, while the student reads it as $55^\circ$, leading to a
different answer. The complete teacher target can therefore encode visually
grounded corrections unsupported by the student's local response geometry,
as depicted in Figure~\ref{fig:motivation_overview}(b).

We expose this mismatch by controlling how far the distillation target moves
from the student distribution toward the teacher distribution. An
interpolation strength of $\alpha=1$ recovers the complete
teacher target, whereas smaller values retain only a fraction of the same
teacher-directed correction. This intervention changes the target strength
while keeping the data, model pair, training schedule, and OPD objective
fixed. We then measure the fraction of the prescribed KL gap removed by the
student after one epoch. As shown in
Figure~\ref{fig:motivation_overview}(c), an 8B-to-2B student removes $56.3\%$
of the gap at $\alpha=0.5$, but only $46.6\%$ under the complete teacher
target. The same intermediate target also improves MathVista Geometry from
$55.29$ to $57.03$. Thus, increasing the target all the way to the teacher
distribution makes supervision less fitable and degrades downstream
reasoning.

This evidence calls for a capacity-aware OPD target. Uniformly weakening
teacher supervision is insufficient because it scales both realizable and
unrealizable corrections. Instead, the target should preserve the component
of the teacher correction supported by the student's visual response
geometry. The appropriate geometry follows from the distillation objective
itself: locally, KL divergence induces the Fisher metric over predictive
distributions \citep{amari1998natural}. Fisher geometry therefore provides a
principled criterion for extracting the teacher-directed change that the
student can express through its current visual pathway.

We introduce \textbf{Fisher-Projected On-Policy Distillation (FP-OPD)}.
FP-OPD expresses teacher supervision as a correction to the student's
log-probabilities, estimates the student's local visual response space, and
projects the teacher correction onto this space under the Fisher metric. The
projected correction defines a capacity-aware target $q^R$, shown
schematically in Figure~\ref{fig:motivation_overview}(b). FP-OPD then
optimizes the same token-level reverse KL as standard OPD, replacing only the
complete teacher distribution with $q^R$. It therefore retains the on-policy
training protocol and distribution-matching objective of OPD while aligning
the distillation target with the student's visual capacity.

Our contributions are threefold:
\begin{itemize}
    \item We identify a mismatch between student capacity and the
    distillation target in VLMs. The complete teacher distribution can encode
    visual corrections outside the
    response geometry of a compact student. A controlled target-interpolation
    experiment shows that the complete target is both harder to fit and worse
    for downstream reasoning.
    \item We propose FP-OPD, which uses Fisher projection to retain the
    student-supported component of the teacher correction and constructs a
    capacity-aware target without changing the OPD loss.
    \item We demonstrate consistent gains across teacher and student scales
    and teacher variants. In 8B-to-2B distillation, FP-OPD improves all seven
    benchmarks and raises the average by 2.77 points over the base student
    and 1.60 points over standard OPD.
\end{itemize}

\section{Related Work}
\label{sec:related_work}

\begin{figure*}[t]
    \centering
    \includegraphics[width=\textwidth]{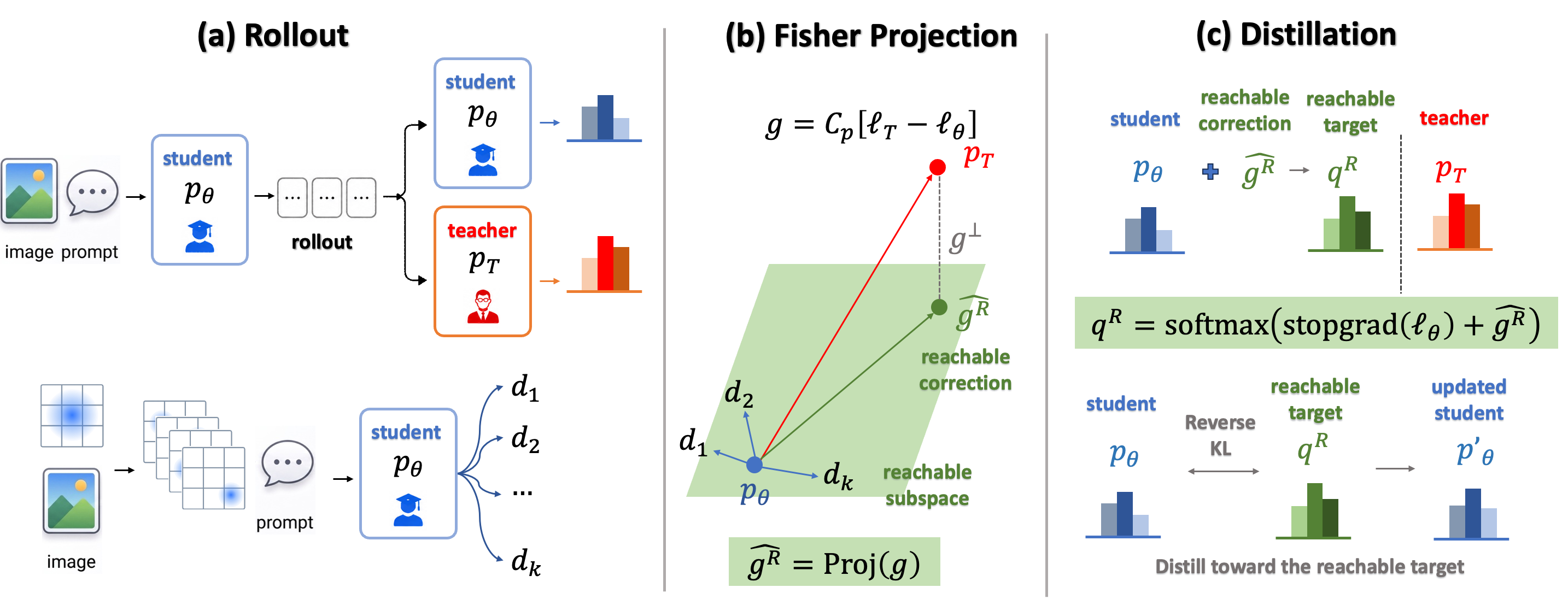}
    \caption{\textbf{Overview of FP-OPD.}
    The student generates an on-policy trajectory, after which the student and
    teacher distributions are evaluated on the same student-visited prefixes.
    Continuous visual perturbations produce local output-response directions
    whose span approximates the student-supported visual response space.
    FP-OPD projects the centered teacher--student gap onto this space under the
    Fisher metric, constructs the reachable target $q^R$, and optimizes it with
    full-vocabulary reverse KL. 
    % Here,
    % $\mathcal C_p[a]
    % =a-\bigl(\sum_{v\in\mathcal V}p(v)a_v\bigr)\mathbf 1$
    % denotes centering under the distribution $p$.
    }
    \label{fig:method_overview}
\end{figure*}

\subsection{Vision--Language Models}

Recent vision--language models have substantially advanced multimodal
perception and reasoning through stronger visual encoders, native-resolution
processing, and large-scale multimodal pretraining. LLaVA-OneVision unifies
single-image, multi-image, and video understanding within one model
\citep{li2024llava}. Qwen2.5-VL introduces dynamic-resolution processing and
improved spatial--temporal modeling \citep{bai2025qwen2}, while InternVL3
jointly learns multimodal and linguistic capabilities through native
multimodal pretraining \citep{zhu2025internvl3}. More recently, Qwen3-VL
incorporates multi-level visual features and enhanced spatial--temporal
position encoding, achieving strong performance across general understanding
and visual reasoning tasks \citep{bai2025qwen3}. These advances also produce
model families with substantially different visual capacities across scales,
motivating FP-OPD to account for whether a teacher's distributional
corrections are supported by the compact student's visual response space.

\subsection{On-Policy Distillation}

On-policy distillation supervises the student at prefixes sampled from its
own policy. GKD-OPD directly matches teacher and student distributions at
student-visited states \citep{agarwal2024policy}, while PG-OPD derives
token-level learning signals from sampled teacher--student log-probability
ratios \citep{lu2025onpolicydistillation}. MOPD combines multiple specialized
teachers \citep{xiao2026mimo,xu2026deepseek,zeng2026glm,yang2026nemotron},
and OPSD uses the same model under privileged and unprivileged contexts as
teacher and student \citep{zhao2026self}. These methods change the
optimization or source of on-policy supervision, whereas FP-OPD retains
direct full-vocabulary reverse-KL matching and changes the target through
Fisher projection.

Recent work also improves the efficiency and reliability of OPD. Prefix OPD
restricts supervision to early reasoning tokens to reduce rollout and
training cost \citep{zhang2026fast}. Vision-OPD transfers privileged regional
perception to the full-image policy through on-policy self-distillation
\citep{yuan2026vision}. Related analyses show that stronger teachers do not
necessarily yield better students when their reasoning patterns or
distributions are poorly aligned \citep{li2026rethinking}. Unlike these
token-selection and optimization strategies, FP-OPD projects the complete
teacher--student distribution gap onto the output space supported by the
student's visual representation.

\subsection{Fisher Geometry and Local Function Spaces}

The Fisher information characterizes the local geometry of predictive
distributions and underlies natural-gradient and trust-region methods
\citep{amari1998natural,schulman2015trust}. Sobolev training and Jacobian
matching transfer derivative information from a target model
\citep{czarnecki2017sobolev,srinivas2018knowledge}, while low-dimensional
optimization studies suggest that useful model changes often lie in compact
subspaces \citep{li2018measuring,gur2018gradient,zhao2024galore}. FP-OPD
differs by using continuous perturbations of the student's visual
representations to estimate a local output space and using Fisher geometry
to project the OPD correction, rather than to define a parameter update or
match teacher derivatives.

% Method section for FP-OPD v6.
% Required packages: amsmath, amssymb, booktabs, multirow, graphicx

\section{Method}

\label{sec:method}

Figure~\ref{fig:method_overview} summarizes FP-OPD. The student first
generates an on-policy trajectory, after which both models are evaluated on
the same student-visited prefixes. At each prefix, FP-OPD expresses the
teacher target as a correction to the student distribution, estimates the
student's local visual response space with detached perturbation probes, and
projects the correction onto this space under the Fisher metric. The
projected correction defines a reachable target $q^R$, which replaces the
complete teacher distribution in the standard full-vocabulary reverse-KL
objective.

\subsection{From OPD to capacity-aware distillation}
\label{sec:from_opd}

Let $x=(I,u)$ denote an image--prompt pair, and let $y_{<t}$ be a prefix
visited by the current student.  At this prefix, the student and teacher
define full-vocabulary distributions
\begin{align}
    p_{\theta,t}
    &=p_{\theta}(\cdot\mid x,y_{<t}), \\
    p_{T,t}
    &=p_T(\cdot\mid x,y_{<t}),
\end{align}
with log-probabilities $\ell_{\theta,t}=\log p_{\theta,t}$ and
$\ell_{T,t}=\log p_{T,t}$. As shown on the left of
Figure~\ref{fig:method_overview}, the student generates the trajectory and
both models are evaluated on the resulting prefixes. Standard on-policy
distillation (OPD) minimizes the token-level reverse KL on these prefixes:
\begin{equation}
    \mathcal L_{\mathrm{OPD}}(x,y)
    =
    \frac{1}{|\mathcal T_y|}
    \sum_{t\in\mathcal T_y}
    \operatorname{KL}\!\left(p_{\theta,t}\,\Vert\,p_{T,t}\right),
    \label{eq:opd_loss}
\end{equation}
where $\mathcal T_y$ is the set of valid completion positions.  The
on-policy sequence determines which prefixes are used for distillation. The
loss itself remains a direct distribution-matching objective over the full
vocabulary.

To express the displacement between these two distributions, consider one
position and omit the subscript $t$. For a categorical distribution $p$ over
vocabulary
$\mathcal V$, define the Fisher inner product and centering operator
\begin{align}
    \langle a,b\rangle_p
    &=\sum_{v\in\mathcal V}p(v)a_vb_v, \\
    \mathcal C_p[a]
    &=a-\left(\sum_{v\in\mathcal V}p(v)a_v\right)\mathbf 1
      =a-\langle a,\mathbf 1\rangle_p\mathbf 1.
    \label{eq:fisher_centering}
\end{align}
Thus, $\mathcal C_p[a]$ subtracts the $p$-weighted mean of $a$ from every
vocabulary coordinate, ensuring
$\langle\mathcal C_p[a],\mathbf 1\rangle_p=0$.
With $p=p_\theta$, the centered teacher--student log-probability gap is
\begin{equation}
    g
    =\mathcal C_p\!\left[\ell_T-\ell_\theta\right].
    \label{eq:teacher_student_gap}
\end{equation}
Because softmax is invariant to an additive constant, this centering does
not change the distribution represented by
$\ell_\theta+(\ell_T-\ell_\theta)=\ell_T$. Consequently, the OPD teacher
target can be written exactly as
\begin{equation}
    p_T
    =\operatorname{softmax}\!\left(\ell_\theta+g\right).
    \label{eq:teacher_as_gap}
\end{equation}
Thus, standard OPD asks the student to follow the complete correction $g$.

For a compact vision--language student, $g$ contains both corrections
supported by its visual representation and corrections that require
teacher-only visual capacity. FP-OPD retains the student-supported component
to construct a capacity-aware OPD target.

\subsection{Local visual reachability}
\label{sec:visual_reachability}

Let the student vision encoder map image $I$ to merged visual embeddings
$Z\in\mathbf{R}^{N\times d}$, and let $\mathcal V_Z$ denote its local visual
representation perturbation space around $Z$. Mapping this space through the
student Jacobian defines the locally reachable output space
\begin{equation}
    \mathcal T_\theta(x,y_{<t})
    =
    \overline{\operatorname{range}}\!\left(
      \left.\mathcal C_p\circ J_Z\ell_\theta\right|_{\mathcal V_Z}
    \right).
    \label{eq:ideal_tangent_space}
\end{equation}
This space contains the first-order distributional changes supported by the
student's current visual state. It is a property of the student and context,
independent of the finite probes used to estimate it.

\subsection{Finite-difference tangent approximation}
\label{sec:finite_difference_tangent}

Explicitly constructing $\mathcal T_\theta$ is intractable. We reshape the
tokens to an $H\times W$ grid and choose $K$ spatial weighting fields
$\mathcal U_K=\{m_1,\ldots,m_K\}$, where
$m_k\in[0,1]^{H\times W}$. With per-image mean $\bar Z$, define
\begin{equation}
    \begin{aligned}
    \delta Z_k&=m_k\odot(\bar Z-Z),\qquad m_k\in\mathcal U_K,\\
    Z^{(k)}&=Z+\varepsilon_{\mathrm{fd}}\delta Z_k.
    \end{aligned}
    \label{eq:continuous_perturbation}
\end{equation}
Binary or soft fields encode spatial patterns and perturb merged and
intermediate features without deleting tokens. The scalar
$\varepsilon_{\mathrm{fd}}$ controls only the interpolation magnitude.

Let $\ell_\theta^{(k)}$ be the student log-probabilities obtained with
$Z^{(k)}$. One detached forward gives the finite-difference response
\begin{equation}
    d_k
    =\mathcal C_p\!\left[
      \frac{\ell_\theta-\ell_\theta^{(k)}}{\varepsilon_{\mathrm{fd}}}
    \right],
    \qquad k=1,\ldots,K.
    \label{eq:tangent_direction}
\end{equation}
The empirical tangent space is
\begin{equation}
    D_K=[d_1,\ldots,d_K],
    \qquad
    \widehat{\mathcal T}_{\theta,K}
    =\operatorname{span}(D_K)
    \approx\mathcal T_\theta.
    \label{eq:empirical_tangent_space}
\end{equation}
\noindent\textbf{Probe procedure.}
\textsc{BuildBasis}$(I,\mathcal U_K)$: (1) run the vision tower once to obtain
$Z$. (2) Instantiate the $K$ spatial fields for $(H,W)$. (3) For every $m_k$,
perturb features, run a detached language-model forward, and compute $d_k$.
(4) Return $D_K=[d_1,\ldots,d_K]$.

\begin{table}[t]
    \centering

    \caption{Main results across three teacher--student settings. ID datasets
    measure visual mathematical reasoning close to Geo3K, whereas OOD
    datasets measure broader multimodal generalization. Average is the
    unweighted mean over all seven datasets. Bold denotes the highest filled
    value within each student-size block.}
    \label{tab:main_results}

    \footnotesize
    \setlength{\tabcolsep}{1.5pt}
    \renewcommand{\arraystretch}{1.10}

    % c: Student, c: Teacher, l: Method, followed by 8 centered columns.
    % @{} removes unnecessary padding at the left and right edges.
    \begin{tabular}{@{}ccl*{8}{c}@{}}
        \toprule
        \multirow{2}{*}{Student} &
        \multirow{2}{*}{Teacher} &
        \multirow{2}{*}{Method} &
        \multicolumn{4}{c}{ID: Mathematical reasoning} &
        \multicolumn{3}{c}{OOD: General multimodal} &
        \multirow{2}{*}{Average} \\
        \cmidrule(lr){4-7}
        \cmidrule(lr){8-10}

        & & &
        WeMath &
        \shortstack{Math\\Vista} &
        \shortstack{Math\\Verse} &
        \shortstack{Math\\Vision} &
        MMMU &
        Hallusion &
        MMStar &
        \\

        \midrule

        \multirow[c]{6}{*}{\shortstack{Qwen3-VL-2B\\Instruct}}
        & -- & Base
        & 52.51 & 53.20 & 47.60 & 18.29
        & 43.90 & 67.31 & 56.70 & 48.50 \\

        & -- & GRPO
        & 60.09 & 57.15 & 49.95 & 20.80
        & 45.14 & 67.27 & 58.41 & 51.26 \\

        & 8B Instruct & OPD
        & 56.16 & 60.30 & 46.80 & 18.25
        & 41.72 & 68.01 & 56.46 & 49.67 \\

        & \shortstack{8B Instruct-\\[-1pt]GRPO} & OPD
        & 59.78 & 62.05 & 49.76 & 21.04
        & 45.43 & 68.17 & 58.14 & 52.05 \\

        \cmidrule(lr){2-11}

        & 8B Instruct & \textbf{FP-OPD}
        & \scoregain{58.71}{+2.55}
        & \scoregain{61.44}{+1.14}
        & \scoregain{48.04}{+1.24}
        & \scoregain{20.49}{+2.24}
        & \scoregain{44.67}{+2.95}
        & \scoregain{68.18}{+0.17}
        & \scoregain{57.38}{+0.92}
        & \scoregain{51.27}{+1.60} \\

        & \shortstack{8B Instruct-\\[-1pt]GRPO}
        & \textbf{FP-OPD}
        & \scoregain{\mathbf{60.32}}{+0.54}
        & \scoregain{\mathbf{63.19}}{+1.14}
        & \scoregain{\mathbf{50.90}}{+1.14}
        & \scoregain{\mathbf{21.64}}{+0.60}
        & \scoregain{\mathbf{46.08}}{+0.65}
        & \scoregain{\mathbf{68.47}}{+0.30}
        & \scoregain{\mathbf{58.85}}{+0.71}
        & \scoregain{\mathbf{52.78}}{+0.73} \\

        \midrule

        \multirow[c]{3}{*}{\shortstack{Qwen3-VL-8B\\Instruct}}
        & -- & Base
        & 69.58 & 72.78 & 61.45 & 33.00
        & 58.46 & \textbf{75.07} & 67.88 & 62.60 \\

        & 32B Instruct & OPD
        & 70.16 & 73.79 & 61.87 & 35.43
        & 58.42 & 74.92 & \textbf{68.24} & 63.26 \\

        \cmidrule(lr){2-11}

        & 32B Instruct & \textbf{FP-OPD}
        & \scoregain{\mathbf{72.39}}{+2.23}
        & \scoregain{\mathbf{74.60}}{+0.81}
        & \scoregain{\mathbf{62.92}}{+1.05}
        & \scoregain{\mathbf{36.40}}{+0.97}
        & \scoregain{\mathbf{59.74}}{+1.32}
        & \scoregain{74.39}{-0.53}
        & \scoregain{68.03}{-0.21}
        & \scoregain{\mathbf{64.06}}{+0.80} \\

        \bottomrule
    \end{tabular}
\end{table}

\subsection{Fisher projection of the OPD correction}
\label{sec:fisher_projection}

The projection metric should be consistent with the divergence optimized by
OPD.  For a centered score displacement $\delta$, the local expansion of
reverse KL is
\begin{equation}
    \operatorname{KL}\!\left(
      p\,\Vert\,
      \operatorname{softmax}(\log p+\delta)
    \right)
    =\frac{1}{2}\langle\delta,\delta\rangle_p
     +o(\|\delta\|^2).
    \label{eq:local_kl_geometry}
\end{equation}
The Fisher inner product in Eq.~\eqref{eq:fisher_centering} is therefore the
local geometry induced by the OPD loss itself.  We use this geometry to find
the tangent-space component closest to the complete OPD correction, as shown
in the middle of Figure~\ref{fig:method_overview}:
\begin{equation}
    \widehat g^{\,R}
    =\arg\min_{u\in\widehat{\mathcal T}_{\theta,K}}
      \langle g-u,g-u\rangle_p.
    \label{eq:fisher_projection_objective}
\end{equation}

For each sample and completion position, define
\begin{align}
    A_{ij}&=\langle d_i,d_j\rangle_p, \\
    b_i&=\langle d_i,g\rangle_p.
    \label{eq:gram_cross}
\end{align}
$A$ is the Fisher Gram matrix of the probes, and $b$ aligns them with $g$.
Because finite-difference directions may be correlated or nearly degenerate,
we solve the ridge-stabilized system
\begin{align}
    \lambda
    &=\rho\,\frac{1}{K}\operatorname{tr}(A)+\epsilon_{\mathrm{num}},
    \qquad \rho=10^{-4}, \\
    c&=(A+\lambda I)^{-1}b,
    \label{eq:ridge_solve}
\end{align}
and obtain
\begin{align}
    \widehat g^{\,R}
    &=D_Kc=\sum_{k=1}^{K}c_kd_k, \\
    g^\perp&=g-\widehat g^{\,R}.
    \label{eq:projected_gap}
\end{align}
The position-wise solve adapts $\widehat g^{\,R}$ to the current visual and
linguistic context. The residual $g^\perp$ lies outside the local probe space
and is excluded from the current target.

\subsection{Fisher-projected OPD objective}
\label{sec:fp_opd_objective}

Standard OPD constructs its target by adding the complete gap $g$ to the
student scores, as shown in Eq.~\eqref{eq:teacher_as_gap}.  FP-OPD makes one
change: it replaces $g$ with its Fisher-projected component
$\widehat g^{\,R}$. As shown in the upper-right part of
Figure~\ref{fig:method_overview}, the resulting target is
\begin{equation}
    q^R
    =\operatorname{softmax}\!\left(
      \operatorname{stopgrad}(\ell_\theta)
      +\widehat g^{\,R}
    \right).
    \label{eq:reachable_target}
\end{equation}
The detached clean logits anchor the current student distribution, and
$\widehat g^{\,R}$ moves this anchor only within the estimated visual tangent
space. Thus, $q^R$ is fixed while the clean student distribution is updated.
If $g$ lies entirely in the empirical tangent space, then
$\widehat g^{\,R}=g$ and $q^R=p_T$, recovering standard OPD.  If $g$ is
orthogonal to that space, then $\widehat g^{\,R}=0$ and $q^R$ reduces to the
detached student distribution, producing no update along the unsupported
direction.

After constructing $q^R$, FP-OPD uses exactly the same token-level reverse-KL
form as Eq.~\eqref{eq:opd_loss}, as shown in the lower-right part of
Figure~\ref{fig:method_overview}:
\begin{equation}
    \mathcal L_{\mathrm{FP\text{-}OPD}}(x,y)
    =
    \frac{1}{|\mathcal T_y|}
    \sum_{t\in\mathcal T_y}
    \operatorname{KL}\!\left(p_{\theta,t}\,\Vert\,q^R_t\right).
    \label{eq:fp_opd_loss}
\end{equation}
Thus, FP-OPD is OPD with a capacity-aware target:
\begin{equation}
    \begin{gathered}
        \underbrace{p_T
          =\operatorname{softmax}(\ell_\theta+g)}_{\text{OPD target}}
        \\[-0.2em]
        \Downarrow
        \\[-0.2em]
        \underbrace{q^R
          =\operatorname{softmax}\!\left(
          \operatorname{stopgrad}(\ell_\theta)+\widehat g^{\,R}
          \right)}_{\text{FP-OPD target}}.
    \end{gathered}
    \label{eq:opd_to_fp_opd}
\end{equation}
FP-OPD changes only the target while retaining reverse KL over the full
vocabulary and the same token reduction. Gradients flow through the clean
student distribution, while the teacher, probe responses, projection, and
$q^R$ remain detached. The student-generated sequence selects the visited
prefixes, with no policy-gradient advantage, reward weighting, or importance
ratio.

\begin{table*}[t]
    \centering
    \footnotesize
    \renewcommand{\arraystretch}{1.08}
    \begin{tabular*}{\textwidth}{@{\extracolsep{\fill}}l*{7}{c}@{}}
        \toprule
        \multirow{2}{*}{Benchmark}
        & Base
        & \multicolumn{2}{c}{Standard On-Policy Distillation}
        & \multicolumn{2}{c}{FP-OPD (ours)}
        & \multicolumn{2}{c}{Improvement} \\
        & \begin{tabular}[c]{@{}c@{}}
              Acc@8\\
              \ensuremath{(T=1.0)}
          \end{tabular}
        & \begin{tabular}[c]{@{}c@{}}
              Acc@1\\
              (greedy)
          \end{tabular}
        & \begin{tabular}[c]{@{}c@{}}
              Acc@8\\
              \ensuremath{(T=1.0)}
          \end{tabular}
        & \begin{tabular}[c]{@{}c@{}}
              Acc@1\\
              (greedy)
          \end{tabular}
        & \begin{tabular}[c]{@{}c@{}}
              Acc@8\\
              \ensuremath{(T=1.0)}
          \end{tabular}
        & \begin{tabular}[c]{@{}c@{}}
              Acc@1\\
              (greedy)
          \end{tabular}
        & \begin{tabular}[c]{@{}c@{}}
              Acc@8\\
              \ensuremath{(T=1.0)}
          \end{tabular} \\

        \midrule
        \multicolumn{8}{c}{
            \textbf{Qwen3-VL-32B-Instruct
            \ensuremath{\rightarrow}
            Qwen3-VL-8B-Instruct}
        } \\
        \midrule

        MMMU
        & 58.46
        & 58.33
        & 58.42
        & \textbf{60.00}
        & \textbf{59.74}
        & \textbf{+1.67}
        & \textbf{+1.32} \\

        WeMath
        & 69.58
        & 68.97
        & 70.16
        & \textbf{71.90}
        & \textbf{72.39}
        & \textbf{+2.93}
        & \textbf{+2.23} \\

        MathVista
        & 72.78
        & 74.00
        & 73.79
        & \textbf{74.60}
        & \textbf{74.60}
        & \textbf{+0.60}
        & \textbf{+0.81} \\

        MathVerse
        & 61.45
        & 61.37
        & 61.87
        & \textbf{62.94}
        & \textbf{62.92}
        & \textbf{+1.57}
        & \textbf{+1.05} \\

        MathVision
        & 33.00
        & 35.16
        & 35.43
        & \textbf{36.25}
        & \textbf{36.40}
        & \textbf{+1.09}
        & \textbf{+0.97} \\

        HallusionBench
        & \textbf{75.07}
        & \textbf{74.67}
        & \textbf{74.92}
        & 74.40
        & 74.39
        & -0.27
        & -0.53 \\

        MMStar
        & 67.88
        & 67.33
        & \textbf{68.24}
        & \textbf{67.53}
        & 68.03
        & \textbf{+0.20}
        & -0.21 \\

        \midrule
        \textbf{Average}
        & 62.60
        & 62.83
        & 63.26
        & \textbf{63.95}
        & \textbf{64.06}
        & \textbf{+1.12}
        & \textbf{+0.80} \\

        \bottomrule
    \end{tabular*}

    \caption{Greedy and sampled results for the
    Qwen3-VL-32B-Instruct \ensuremath{\rightarrow}
    Qwen3-VL-8B-Instruct transfer setting. Improvement is FP-OPD minus
    Standard OPD under the same decoding setting.}
    \label{tab:opd_fp_opd_improvement}
\end{table*}

\begin{table*}[t]
    \centering
    \small
    \setlength{\tabcolsep}{3.2pt}
    \renewcommand{\arraystretch}{1.08}
    \resizebox{0.99\textwidth}{!}{%
    \begin{tabular}{llcccccccc}
        \toprule
        Factor & Variant & WeMath & MathVista & MathVerse &
        MathVision & MMMU & Hallusion & MMStar & Average \\
        \midrule
        \multirow{3}{*}{FD step}
        & $0.025$
        & 57.51 & 60.86 & \textbf{48.07} & 19.84
        & \textbf{44.82} & \textbf{68.93} & \textbf{57.84} & 51.12 \\
        & $0.05$
        & \textbf{58.71} & \textbf{61.44} & 48.04 & \textbf{20.49}
        & 44.67 & 68.18 & 57.38 & \textbf{51.27} \\
        & $0.10$
        & 57.44 & 60.86 & 48.00 & 19.91
        & 44.57 & 67.93 & 57.30 & 50.86 \\
        \midrule
        \multirow{3}{*}{No. of tangent probes}
        & $K=2$
        & 58.18 & 60.75 & 47.88 & 20.33
        & 44.75 & \textbf{68.51} & 57.69 & 51.16 \\
        & $K=4$
        & \textbf{58.71} & \textbf{61.44} & \textbf{48.04}
        & \textbf{20.49} & 44.67 & 68.18 & 57.38 & \textbf{51.27} \\
        & $K=8$
        & 58.43 & 60.31 & 47.60 & 20.24
        & \textbf{45.33} & 67.74 & \textbf{58.00} & 51.09 \\
        \midrule
        \multirow{3}{*}{Projection}
        & Euclidean + visual probes
        & 44.01 & 44.83 & 36.62 & 14.67
        & 38.93 & 65.40 & 52.90 & 42.48 \\
        & Fisher + random probes
        & 57.44 & 60.69 & 47.90 & 19.65
        & \textbf{44.85} & \textbf{68.40} & \textbf{57.38} & 50.90 \\
        & Fisher + visual probes
        & \textbf{58.71} & \textbf{61.44} & \textbf{48.04}
        & \textbf{20.49} & 44.67 & 68.18 & \textbf{57.38}
        & \textbf{51.27} \\
        \bottomrule
    \end{tabular}
    }
    \caption{FP-OPD ablations under \textsc{Avg@8}.
    Bold marks the best result within each block.}
    \label{tab:fp_opd_ablations}
\end{table*}

\begin{figure*}[t]
    \centering
    \includegraphics[width=\textwidth]
    {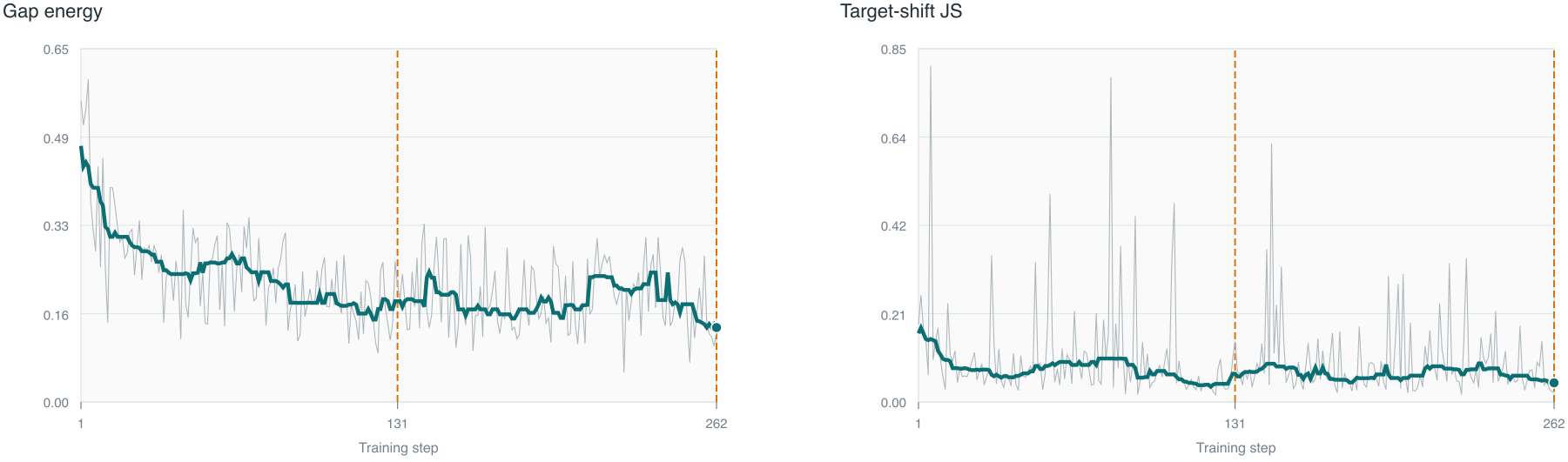}
    \caption{Teacher--student gap energy and target-shift divergence during
    32B-to-8B FP-OPD training.}
    \label{fig:fp_opd_training_dynamics}
\end{figure*}
\section{Experiments}
\label{sec:experiments}

\subsection{Experimental Setup}
\label{sec:experimental_setup}

\paragraph{Models and data.}
We study three Qwen3-VL transfer settings. The first two use
Qwen3-VL-2B-Instruct as the student and either Qwen3-VL-8B-Instruct or its
Geo3K-GRPO variant as the teacher. The third setting distills
Qwen3-VL-32B-Instruct into Qwen3-VL-8B-Instruct. All distillation experiments
use Geo3K \citep{lu2021inter} as the training corpus.

\paragraph{Training.}
We train for two epochs using a step-wise linear learning-rate decay with no
warmup.
% The learning rate decreases from $1\times10^{-6}$ to
% $6\times10^{-7}$ during training.
We use BF16 precision, one example per
GPU, gradient accumulation of 8, and four on-policy rollouts per prompt. The
maximum sequence and completion lengths are 4,096 and 2,048 tokens,
respectively.
% We use the full vocabulary, $\beta=1$, distillation
% temperature 1.0, and gradient clipping at 1.0.
FP-OPD uses four fixed, unsampled binary fields selecting the top, bottom,
left, and right boundary bands. Each band spans fraction $\gamma=0.25$ of
the corresponding grid axis. Fields are instantiated on each dynamic image
grid and use finite-difference step $\varepsilon_{\mathrm{fd}}=0.05$.
Checkpoints are saved once per epoch.
Complete training, method, and evaluation
configurations are provided in Appendix~\ref{app:full_training_configuration}.

\paragraph{Baselines and benchmarks.}
We compare against the base student, direct GRPO post-training, and Standard
OPD. Evaluation covers four mathematical-reasoning benchmarks: WeMath
\citep{qiao2025we}, MathVista \citep{lu2024mathvista}, MathVerse
\citep{zhang2024mathverse}, and MathVision \citep{wang2024measuring}.
We additionally evaluate three general multimodal benchmarks:
MMMU \citep{yue2024mmmu}, HallusionBench
\citep{guan2024hallusionbench}, and MMStar \citep{chen2024we}. Unless
explicitly marked, scores are \textsc{Avg@8} with temperature 1.0,
top-$p=0.95$, and at most 2,048 generated tokens.

\subsection{Main Results}
\label{sec:main_results}

\paragraph{8B-to-2B distillation.}
Table~\ref{tab:main_results} first compares the two settings with a 2B
student. With the 8B-Instruct teacher, FP-OPD reaches an average of 51.27,
improving the base student by 2.77 points and Standard OPD by 1.60 points.
It exceeds both baselines on all seven benchmarks, demonstrating consistent
gains across in-domain mathematical reasoning and out-of-domain multimodal
evaluation. Relative to Standard OPD, the largest improvements occur on
MMMU (+2.95), WeMath (+2.55), and MathVision (+2.24), while MathVista
(+1.14), MathVerse (+1.24), HallusionBench (+0.17), and MMStar (+0.92) also
improve. This broad improvement shows that FP-OPD strengthens mathematical
reasoning without sacrificing the student's general multimodal performance.

FP-OPD benefits further from the stronger Geo3K-GRPO teacher. It obtains
52.78 on average and achieves the best result in every column of the
2B-student block. This is 1.51 points above FP-OPD with the 8B-Instruct
teacher, 1.52 points above direct GRPO post-training of the student, and 0.73
points above Standard OPD with the same teacher. Compared with direct GRPO,
FP-OPD improves all seven benchmarks, including +6.04 on MathVista, +1.20 on
HallusionBench, and +0.94 on MMMU. These results show that FP-OPD can exploit
a stronger reasoning teacher while maintaining gains across both ID and OOD
evaluation.

\paragraph{Scaling to larger models.}
Table~\ref{tab:main_results} also reports the larger 32B-to-8B setting.
FP-OPD reaches an average of 64.06, improving the base 8B student by 1.46
points and Standard OPD by 0.80 points. It raises all four mathematical
benchmarks, with gains over Standard OPD of +2.23 on WeMath, +0.81 on
MathVista, +1.05 on MathVerse, and +0.97 on MathVision. MMMU also improves
by +1.32. Overall, FP-OPD exceeds the base student on six of seven benchmarks
and Standard OPD on five of seven. The two small differences on
HallusionBench (-0.53) and MMStar (-0.21) are outweighed by the consistent
reasoning gains. The improvement from a 32B teacher to an already strong 8B
student shows that FP-OPD remains effective as both model sizes increase.

\paragraph{Decoding robustness.}
Table~\ref{tab:opd_fp_opd_improvement} shows that FP-OPD remains effective
under deterministic decoding. Its greedy average reaches 63.95, improving
Standard OPD by 1.12 points and winning on six of seven benchmarks. The
largest gains occur on WeMath (+2.93), MMMU (+1.67), and MathVerse (+1.57),
followed by MathVision (+1.09), MathVista (+0.60), and MMStar (+0.20).
FP-OPD also reaches 64.06 under \textsc{Avg@8}, an improvement of 0.80
points. The similar per-benchmark trends under greedy and sampled decoding
show that the benefit of FP-OPD reflects a stronger student model rather
than sampling variance.

\subsection{Ablation Studies}
\label{sec:ablations}

\paragraph{Finite-difference step.}
The default $\varepsilon_{\mathrm{fd}}=0.05$ obtains the best average of
51.27, compared with 51.12 at $0.025$ and 50.86 at $0.10$. The smaller step
is slightly better on MathVerse and the three general multimodal benchmarks,
whereas $0.05$ performs best on WeMath, MathVista, and MathVision. The narrow
spread shows that FP-OPD is stable around its default, while the moderate
perturbation gives the best overall balance.

\paragraph{Number of tangent probes.}
Four probes achieve the highest average of 51.27 and lead all four
mathematical-reasoning benchmarks. The settings with two and eight probes
remain competitive at 51.16 and 51.09, respectively, with isolated gains on
the general multimodal benchmarks. The $K=2$ variant uses centered horizontal
and vertical bands, $K=4$ uses the top, bottom, left, and right boundary
bands defined in the training setup, and $K=8$ adds four corner regions.
Performance is not monotonic, so we use $K=4$ by default.

\paragraph{Projection geometry and probe construction.}
Replacing the Fisher metric with Euclidean projection sharply reduces the
average from 51.27 to 42.48, a drop of 8.79 points. This result confirms that
the projection geometry must be consistent with the local geometry induced
by the KL distillation objective. Keeping the Fisher metric but replacing
the fixed boundary-band probes with random probes yields 50.90. The
smaller 0.37-point decrease shows that Fisher-consistent projection accounts
for most of the benefit, while spatially structured probes provide an
additional, systematic improvement. We therefore combine the Fisher metric
with a four-probe tangent approximation in FP-OPD.

\subsection{Mechanism Analysis}
\label{sec:mechanism_analysis}

\textbf{Evolution of the projected correction.}
We inspect the training diagnostics over two epochs of the 32B-to-8B
full-Geo3K run. Step 131 and step 262 in Figure~\ref{fig:fp_opd_training_dynamics} mark the
ends of epochs 1 and 2, respectively. The figure reports two complementary
quantities. \emph{Gap energy},
$E_{\mathrm{gap}}=\langle g,g\rangle_{p_\theta}$, is the squared Fisher norm
of the complete teacher--student log-probability correction before
projection. It measures the local distributional discrepancy that Standard
OPD would ask the student to match. \emph{Target-shift JS},
$\operatorname{JS}(p_\theta\|q^R)$, measures the actual displacement from
the current student distribution to the Fisher-projected target after
projection and normalization. Thus, the first metric quantifies how far the
teacher is from the student, whereas the second quantifies how far FP-OPD
actually moves its target.

Both quantities decrease rapidly during the first epoch and remain low
during the second. Comparing the first and last 32 updates, median gap energy
falls from 0.299 to 0.169 ($-43.3\%$), while median target-shift
Jensen--Shannon divergence falls from 0.0987 to 0.0634 ($-35.8\%$). The
former shows that the predictive mismatch on student-visited prefixes
shrinks during training. The latter shows that the reachable target $q^R$
becomes progressively closer to the student's current predictive
neighborhood. Their joint decline indicates that the student is absorbing
the projected supervision rather than repeatedly receiving an equally large
correction. Neither signal collapses immediately, indicating continued
learning rather than an inactive distillation target.

% Conclusion for FP-OPD v6.

\section{Conclusion}
\label{sec:conclusion}

On-policy distillation aligns supervision with student-visited prefixes, but
still assumes that the complete teacher distribution is suitable for a
compact student. We show that part of the teacher correction lies outside
the student's local visual response space. Our target-interpolation study
confirms that the complete target is harder to fit and yields worse
downstream performance.

We therefore propose Fisher-Projected On-Policy Distillation (FP-OPD), which
projects the teacher correction onto a perturbation-estimated visual tangent
space under the Fisher metric. The resulting reachable target preserves
standard OPD training while adapting supervision to the student's capacity.
FP-OPD improves the seven-benchmark average across teacher and student
scales, with gains that persist under greedy decoding. Overall, FP-OPD makes
teacher supervision locally realizable, enabling compact students to learn
the teacher distribution more effectively across multimodal reasoning tasks.
Future work will explore adaptive tangent-space estimation and more efficient
projection strategies to reduce the additional forward-pass cost.

\newpage
\bibliographystyle{plainnat}
\bibliography{reference}
\end{document}